\documentclass[10pt,twocolumn,letterpaper]{article}

\usepackage{cvpr}

\usepackage{array}
\usepackage[export]{adjustbox} %
\usepackage{times}
\usepackage{epsfig}
\usepackage{graphicx}
\usepackage{amsmath}
\usepackage{amssymb}
\usepackage{graphics}
\usepackage{pifont}
\usepackage{color}
\usepackage{booktabs}
\usepackage{multirow}  %
\usepackage{subcaption}
\usepackage{gensymb}
\usepackage{bm}
\usepackage[table]{xcolor}
\usepackage{cite} %
\usepackage{subcaption} %
\usepackage{microtype}

\graphicspath{{..}, {.}, {figures/}}

\definecolor{citecolor}{RGB}{34,139,34}
\usepackage[pagebackref=true,breaklinks=true,letterpaper=true,colorlinks, citecolor=citecolor,bookmarks=false]{hyperref}

\cvprfinalcopy %

\def\cvprPaperID{4890} %

\definecolor{julieta_colour}{RGB}{117,112,179} %

\begin{document}

\title{Learning to Localize Through Compressed Binary Maps}

\author{%
  Xinkai Wei$^{1,2}\thanks{Equal contribution}$ \quad Ioan Andrei
  B\^{a}rsan$^{1,3}$\footnotemark[1] \quad
  Shenlong Wang$^{1, 3}$\footnotemark[1]  \\
  Julieta Martinez$^{1}$
  \quad Raquel Urtasun$^{1,3}$\\
  $^{1}$Uber Advanced Technologies Group \quad $^{2}${University of Waterloo}
  \quad $^{3}$University of Toronto\\
 \small\texttt{\{xinkai.wei,andreib,slwang,julieta,urtasun\}@uber.com}
}

\newcommand{\bx}{\mathbf{x}}
\newcommand{\bb}{\mathbf{b}}
\newcommand{\ba}{\mathbf{a}}
\newcommand{\bo}{\mathbf{o}}
\newcommand{\bz}{\mathbf{z}}
\newcommand{\bp}{\mathbf{p}}
\newcommand{\bn}{\mathbf{n}}
\newcommand{\bw}{\mathbf{w}}
\newcommand{\cI}{\mathcal{I}}
\newcommand{\cF}{\mathcal{F}}
\newcommand{\cL}{\mathcal{L}}
\newcommand{\cG}{\mathcal{G}}
\newcommand{\cM}{\mathcal{M}}
\newcommand{\by}{\mathbf{y}}
\newcommand{\ut}{^{(t)}}
\newcommand{\up}{^{(t-1)}}
\newcommand{\bt}{\mathbf{t}}

\newcommand{\elidar}{E_\textsc{LiDAR}}
\newcommand{\mlidar}{\text{LiDAR}}
\newcommand{\mgps}{\text{GPS}}

\newcommand{\dyn}{\textsc{Dyn}}
\newcommand{\online}{\textsc{o}}
\newcommand{\map}{\textsc{m}}
\newcommand{\mask}{\textsc{mask}}

\newcommand{\shenlong}[1]{\textcolor{black}{#1}}
\newcommand{\todo}[1]{\textcolor{black}{}}

\maketitle

\begin{abstract}
One of the main difficulties of scaling current localization systems to large
environments is the on-board storage required for the  maps.
In this paper we propose to learn to compress the map
representation such that it is optimal for the localization task. As
a consequence, higher compression rates can be achieved without loss of
localization accuracy when compared to standard coding schemes
that optimize for reconstruction, thus ignoring the end task. 
Our experiments show that it is possible to learn a task-specific compression which
reduces storage requirements by two orders of magnitude over general-purpose
codecs such as WebP without sacrificing performance.

\end{abstract}

\section{Introduction}

One of the fundamental tasks in autonomous driving is the ability to localize the self-driving vehicle (SDV) with respect to a geo-referenced map, as this enables routing the vehicle from point A to point B.
Furthermore, high precision localization enables the use of high definition (HD) maps that capture the static parts of the environment.
This map is used by most self driving teams as a component of perception and motion planning modules.

LiDAR-based localization systems are usually employed for precise localization of SDVs \cite{levinson2007map,levinson2010robust,yoneda2014lidar,wolcott2015fast,deep-gil,Javanmardi2018}.
They rely on having an HD map, which contains dense point clouds
\cite{wolcott2014visual,yoneda2014lidar} and/or intensity LiDAR images of the ground~\cite{levinson2007map,levinson2010robust,wolcott2015fast,deep-gil}.
One of the main difficulties of scaling current localization systems to large
environments is the on-board storage required for the HD maps. For instance,
storing a LiDAR intensity map as a 16-bit PNG file would require roughly 900 GB for
a city such as Los Angeles, and over 168 TB for the entire United
States.\footnote{Based on information from the US Bureau of Transportation
Statistics,
assuming that it takes approximately 4 MB to store a 150 m road
segment as a 16-bit PNG single-channel image
({\scriptsize \url{https://www.bts.gov}
}).}
Storing this information onboard the vehicle is infeasible for scalability
past a single city.
Streaming the HD map data on the go makes the system dependent on a reliable broadband connection, which may not always be available.

\begin{figure}[t]
\vspace{-3mm}
\includegraphics[width=0.99\linewidth]{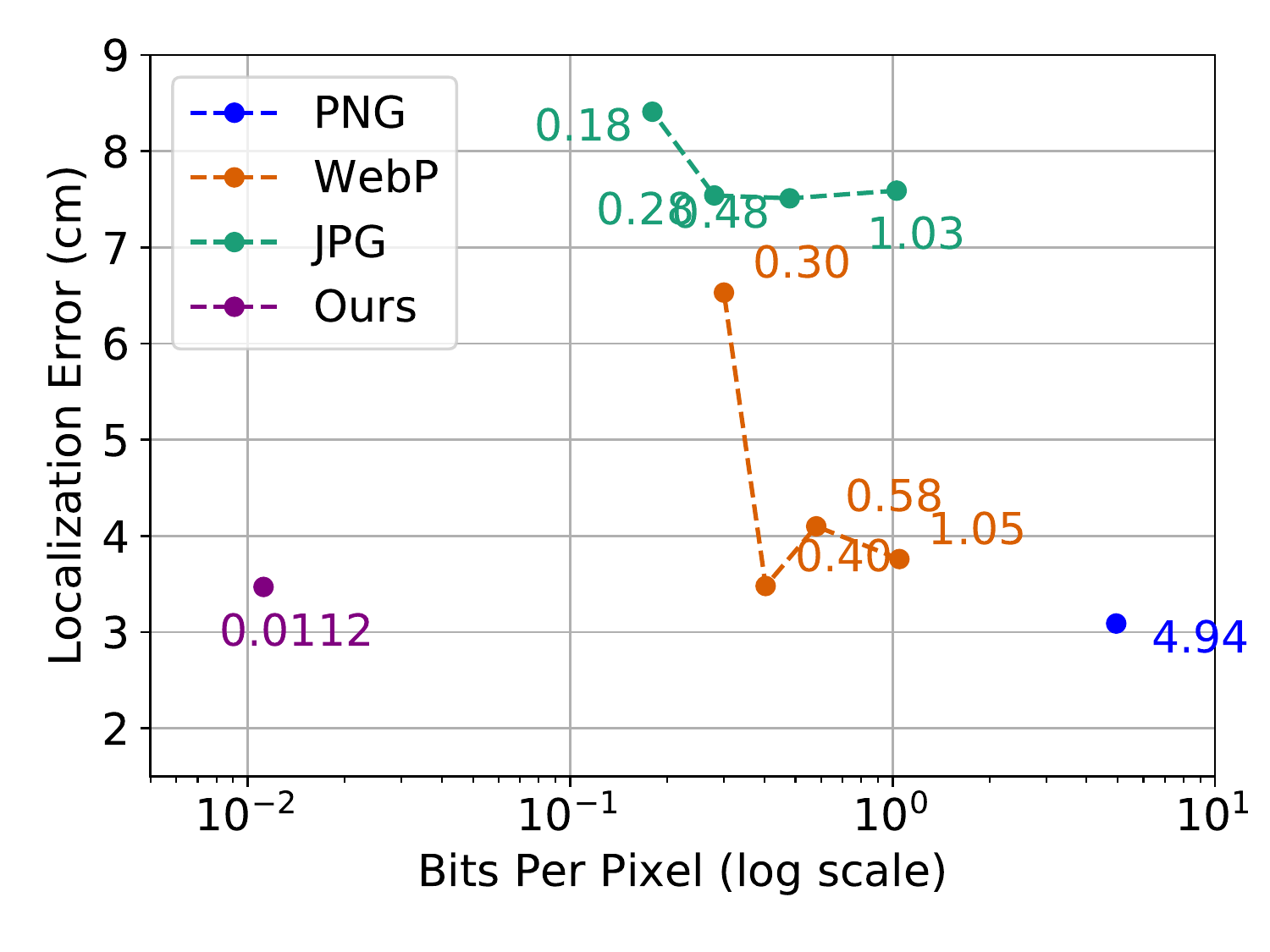}
\vspace{-5mm}
\caption{{\bf End failure rate for localization under different map compression settings.} Lower is better.}
\label{fig:failure_f}
\vspace{-5mm}
\end{figure}
In this paper we propose to learn to compress the map
representation such that it is optimal for the localization task. As
a consequence, higher compression rates can be achieved without loss of
localization accuracy and robustness compared to standard coding schemes
that optimize for reconstruction, thus ignoring the end task. 
In particular, we leverage a fully convolutional network to learn to binarize the
map features, and further compress the binarized representation using run-length encoding on top
of Huffman coding.
Both the binarization net and the decoder are learned end-to-end using a task-specific loss.
We demonstrate the effectiveness of this idea in the context of a 
state-of-the-art LiDAR intensity-based localization system \cite{deep-gil},
and show that it is possible to learn a task-specific compression scheme which
reduces storage requirements by two orders of magnitude over general-purpose
codecs such as WebP, without sacrificing performance.

\begin{figure*}[]
\vspace{-5mm}
\begin{center}
\includegraphics[width=0.88\linewidth]{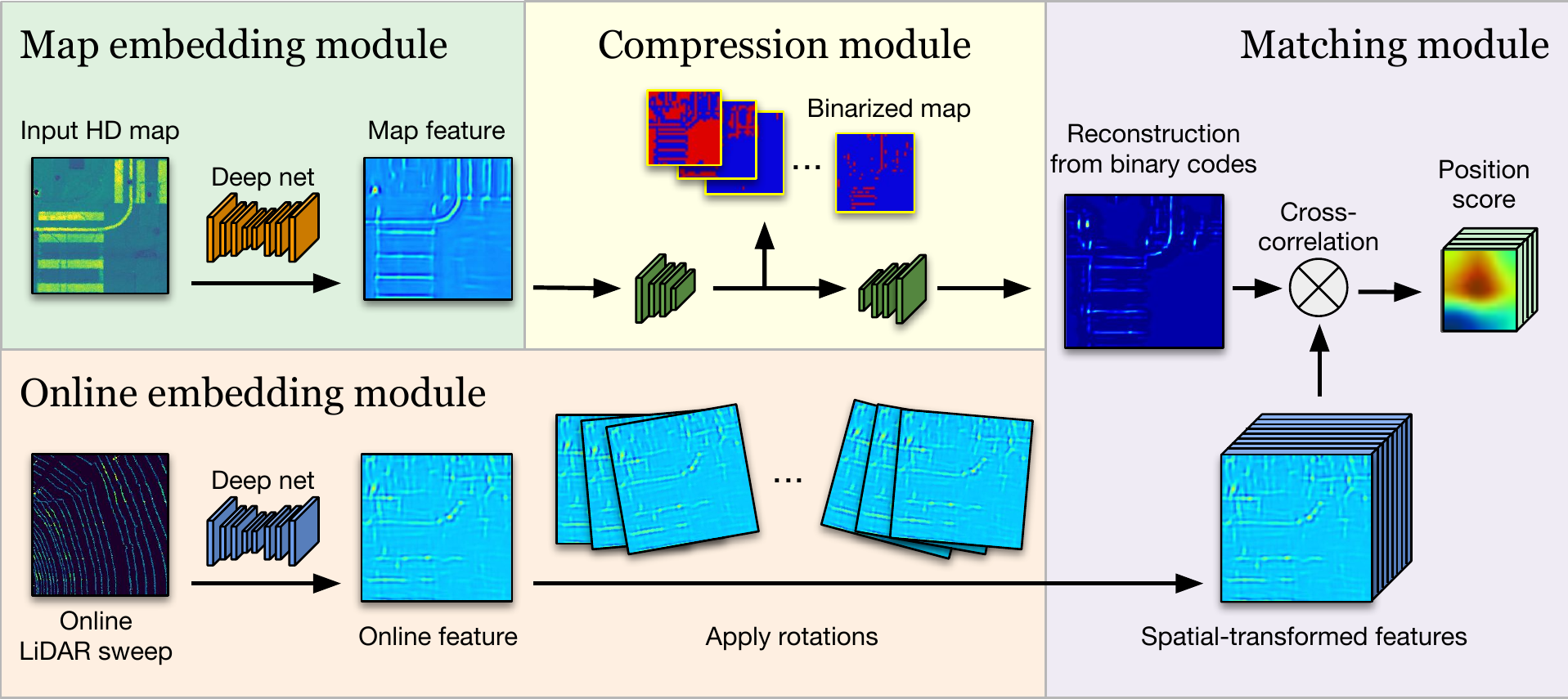} 
\end{center}
\vspace{-5mm}
   \caption{{\bf Architecture overview of our proposed joint compression and localization network.} }
\vspace{-2.5mm}
\label{fig:architecture}
\vspace{-1.5mm}
\end{figure*}

\section{Related Work}

\paragraph{Localization Using HD Maps:}
High-definition maps have been widely used in the field of robot
localization due to their ability to enable centimeter-level accuracy in
a diverse set of
environments, while avoiding some of the computational costs typically
associated with a full SLAM system.
Levinson and Thrun~\cite{levinson2007map} used Graph-SLAM to aggregate LiDAR
observations into a coherent map, which was then used in localization. 
K\"{u}mmerle et al.~\cite{Kummerle2009} use Multi-Level Surface
Maps~\cite{triebel2006multi} with LiDAR, and use the map
for localization and path planning to enable a car to park itself.
Subsequent works have improved the robustness of such approaches by augmenting
the maps with probabilistic occupancy
information~\cite{levinson2010robust,wolcott2015fast}, or by fusing LiDAR
matching results with differential GPS~\cite{Wan2017}.

\paragraph{Lightweight Localization:}
Numerous alternatives to HD maps have been explored over the years
in an attempt to overcome their limitations, such as the dependence on data
collection and offline map construction.
Floros et al.~\cite{Floros13} develop a lightweight extension to visual
odometry which leverages OpenStreetMap to eliminate the drift typically
associated with pure VO. Ma et al.~\cite{ma2016find} extend this idea by using
cues from multiple modalities, such as egocar
trajectory, road type and the position of the sun to localize robustly
within a lightweight map represented as a graph. Recently, Ort et
al.~\cite{Ort2018} used a similar approach, performing road segmentation using
LiDAR to localize within a lightweight topological map with negligible
on-board map storage requirements.
Javanmardi et al.~\cite{Javanmardi2018} extract probabilistic 3D planar
surfaces and 2D vector maps from dense point clouds to create lightweight maps
which are then used in localization, achieving promising results in terms of
accuracy.
In recent years, image-based localization methods~\cite{Brachmann2017,Sattler2017,Radwan2018} have shown  promising
results reaching centimeter-level accuracy in indoor environments.
However, these methods are still not sufficient for centimeter-level
accuracy in outdoor scenarios exhibiting fast motion, such as those occurring
in self-driving. %

\paragraph{Image Compression:}
Image compression is a classical subfield of computer vision and
signal processing. It has seen a great deal of progress in the past few
years thanks to the advent of deep learning. 
In most modern incarnations,
a learning-based compression method consists of an encoder network,
a quantization mechanism (e.g., binary codes), 
and a decoder network which
reconstructs the input from the quantized codes.
Recent
learning-based approaches~\cite{toderici2017full,rippel2017real,mentzer2018conditional} consistently outperform classic compression methods like JPEG2000
and BPG.
While earlier works on learned compression typically used a standard autoencoder
architecture~\cite{Balle_2016}, thereby imposing a fixed
code size for all images, Toderici et al.~\cite{toderici2015variable} overcome this limitation
by using a recurrent neural network as an encoder.
Subsequently,  Toderici et al.~\cite{toderici2017full}
extend the previous results, which were typically presented on resolutions of
$64\times 64$ or less due to performance considerations, showing
state-of-the-art compression rates on full-resolution images. 
Recently, Mentzer et al.~\cite{mentzer2018conditional} proposed a pipeline which
obtained results on par with the state-of-the-art, while also being trainable
end-to-end (encoder, quantizer, decoder).

\paragraph{Learning to Match:}
Learning-based approaches have also been employed in matching problems, which
arise in numerous vision applications, including stereo
matching~\cite{zbontar2015computing,luo2016efficient}, optical
flow~\cite{weinzaepfel2013deepflow,Revaud_2015}, and
map-based localization~\cite{deep-gil}.
In their pioneering work, Zbontar and LeCun~\cite{zbontar2015computing}
proposed modeling the matching cost function used in stereo depth estimation as
a convolutional neural network and learning it from data. Luo et
al.~\cite{luo2016efficient} extend this framework to produce calibrated
probability distributions over the disparities of all pixels, while also
enabling real-time operation through the introduction of an explicit
correlation layer capable of speeding up inference by an order of magnitude
compared to previous work.
Similarly, DeepFlow~\cite{weinzaepfel2013deepflow} applies learning-based
matching to the problem of optical flow estimations. The authors use a learned
matcher to match sparse descriptors, which are then fed into a variational method to
estimate dense flow. EpicFlow~\cite{Revaud_2015} extends this framework by
densifying the sparse matches using a novel interpolation scheme before
performing  variational energy minimization.
The task of map-based localization using matching has also been
approached from a data-driven perspective, using neural networks to learn
representations optimal for matching a LiDAR observation to
a map~\cite{deep-gil}.

\section{End-to-End Compressed Localization}%
\label{sec:method}

LiDAR based localization systems are usually employed by self-driving vehicles to provide  high-precision localization estimates \cite{levinson2010robust, wolcott2015fast, deep-gil}. 
They rely on having an HD map, which contains  dense point clouds \cite{yoneda2014lidar} and/or LiDAR intensity images \cite{levinson2010robust, wolcott2015fast}. 
One of the main difficulties of scaling localization to large environments is the on-board storage required for these maps. 
To tackle this problem, in this paper we propose to learn to compress the map representation such that it is optimal for the localization task. As a consequence, higher compression rates can be achieved without loss of localization accuracy or robustness degradation. In particular,  our approach learns
a compressed deep embedding of the map that can be directly  stored on-board, dramatically reducing the requirements of state-of-the-art LiDAR intensity based localization systems.

In this section, we first revisit the state-of-the-art Deep Ground Intensity Lidar Localizer (Deep GILL)~\cite{deep-gil} and its probabilistic Bayes inference.
We then describe our compression module and show how it can be learned end-to-end jointly with the localizer.

\subsection{Deep GILL Revisit}
Real-time localization with centimeter level accuracy is  critical for most  self-driving cars, as they rely on the semantics captured in HD maps to drive safely. 
In this paper, we follow %
\cite{deep-gil}'s formulation for our localization as a %
recursive Bayes inference problem. 
In particular, the Bayes inference framework combines the LiDAR matching energy, the vehicle dynamics, and the  GPS observations with the  estimates of the previous time step to form the probability of a given location at the current time step. Let  $\mathrm{Bel}(\mathbf{x}_t)$ be   the posterior distribution of the vehicle pose
at time $t$ given all the sensor observations until time step $t$, we have:
\begin{align}
\mathrm{Bel}_t(\bx)  =  \mathrm{Bel}_{t|t-1}(\bx; \mathcal{X}) \cdot P_{\mgps}(\mathcal{G}_t | \bx) \cdot P_{\mlidar}(\cI_t | \bx; \bw),
\label{eq:deepGIL}
\end{align}
where $\bx = \{t_x, t_y, \theta\}$ is the 3-DoF vehicle pose, and $\cI_t \in \mathbb{R}^{N_t\times 4}$ is the online LiDAR sweep containing   $N_t$ points with  geometric  and intensity information; $\mathcal{X}_t={\mathbf{v}_\bx, \mathbf{v}_\theta}$ is the vehicle dynamics  encoding linear  and angular velocity; and $\mathcal{G}_t \in \mathbb{R}^2$ are GPS
observations under Universal
Transverse Mercator (UTM) coordinate system.

\begin{figure*}[t]
\vspace{-3mm}
\begin{center}
\includegraphics[width=0.9\linewidth]{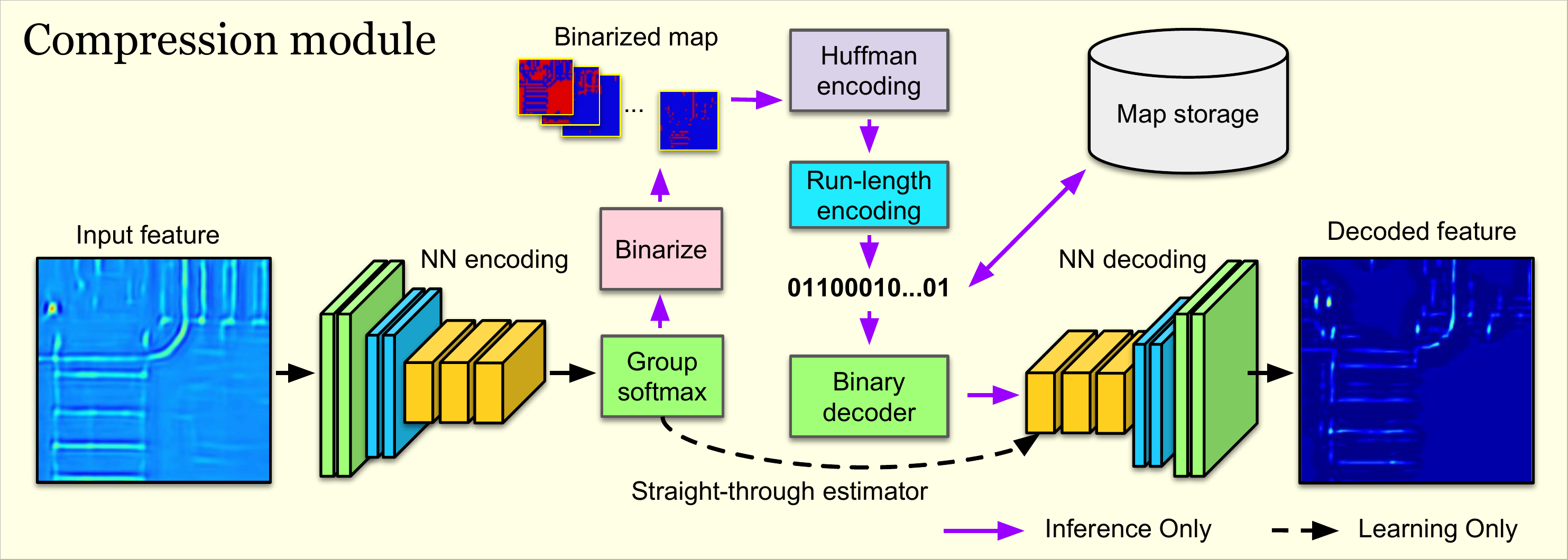} 
\end{center}
\vspace{-5mm}
   \caption{{\bf Our compression module.} We obtain gradients for training with a straight-through estimator.}
\label{fig:compression}
\vspace{-2.5mm}
\end{figure*}

The motion model encodes the fact that the inferred pose should agree with the vehicle dynamics given the previous time step location belief $\mathrm{Bel}_{t-1}(\bx_{t-1}) $, more formally defined as 
\begin{equation}
  \label{eq:motion-model}
  \mathrm{Bel}_{t|t-1} (\bx | \mathcal{X}_t) = \int_{\bx_{t-1}} P(\bx | \mathcal{X}_t, \bx_{t-1}) \mathrm{Bel}_{t-1}(\bx_{t-1}).
\end{equation}
 We use a Gaussian to represent the  vehicle's conditional distribution over vehicle dynamics,
\begin{equation} \label{eq:dynamics-energy}
P(\bx | \mathcal{X}_t, \bx_{t-1})  \propto \mathcal{N} (\bx_{t-1} \oplus \mathcal{X}_t, \Sigma),
\end{equation}
where $\bx_{t-1} \oplus \mathcal{X}_t$ is the last timestamp's pose composed by the current timestamp's velocity observation; $\oplus$ is the pose composition operator, and  $\Sigma$ is the covariance matrix for the velocity estimation. 
The GPS observation model encodes the likelihood of the GPS  as a Gaussian distribution:
\begin{equation}
P_{\mgps} \propto \mathcal{N}([g_x, g_y]^T, \sigma_{\mgps}^2 \mathbf{I} ),
\end{equation}
where $g_x$ and $g_y$ is the map-relative position $\bx$ converted to UTM coordinate system. 

The LiDAR matching model encodes the agreement between the current online LiDAR observation and the map indexed at the hypothesized pose $\bx$:
\begin{equation}
  P_{\mlidar} \propto s \left(\pi\left(f(\cI; \bw_\online), \bx\right), g(\cM; \bw_\map) \right),
\end{equation}
where $f(\cdot)$ and $g(\cdot)$ are the embedding network over online LiDAR sweeps and maps respectively,  and $\pi$ is a rigid transform that converts the online embedding image to the map coordinates using the hypothesized pose $\bx$; $\cM$ is the dense LiDAR intensity map representation, and  $s$ is the correlation operator between the online embedding and map embedding. The computation of this term can be written as a feed-forward network as shown in \cite{deep-gil}. 

While effective for localization, \shenlong{the dense intensity map used in Deep GILL \cite{deep-gil} requires a large amount of on-board storage.
This prevents the method from  scaling to larger operational domains.} To tackle this problem, in this paper we introduce a novel learning-to-compress module that reduces the storage of the map significantly, allowing us to potentially store maps for the full continent.  Next,  we  describe our new compressed model. 

\shenlong{\subsection{Deep Localization with Map Compression}}

Unlike previous compression networks that aim at optimizing the reconstruction error or perceived visual quality, in this paper we argue that optimizing for the tasks at hand is important to further reduce the storage requirements. Towards this goal, we extends the architecture of \cite{deep-gil} and include a compression module responsible for encoding the map with binary codes through deep convolutional neural networks. Importantly, our encoding can be learned end-to-end with our localizer.

We refer the reader to Fig.~\ref{fig:architecture} for an illustration of  the overall architecture of our  joint compression and matching network. 
 Our overall end-to-end network includes three components. First, our embedding module takes  the  map $\mathcal{M}$ and the online LiDAR sweep $\mathcal{I}$ as input, and computes a deep embedding representation of both.  
 A compression module is then applied over the map embedding layer, which converts the high-dimensional float-valued deep embedding map to a compact convolutional binary code representation. This representation is  used as a compact storage of the map. A decoding module is then employed to decode the binary codes back to the real-valued embedding representation. Finally, matching  is conducted between the reconstructed map embedding  and the online embedding. This gives us a score for each possible transformation. We use softmax to build the probability $P_{\mlidar}$ over our localization search space from the raw matching score. We now describe the modules in more detail. 

\paragraph{Embedding Module:} The embedding should capture robust yet discriminative contextual features while preserving pixel-accurate details for precise matching. Motivated by this fact, we designed this module to be a fully convolutional encoder-decoder network following \cite{deep-gil}. \shenlong{It has an U-Net architecture \cite{unet}. The encoder consists of four blocks, each of which has two stride-1 $3\times 3$ conv layers and one stride-2 $3\times3$ conv layer that down-samples the feature map by a factor of 2. The numbers of channel per each block are 64, 128, 256, 512, respectively. The decoder network has four decoder blocks, each of which takes the last decoder block's output and the corresponding encoder layer feature as input in an additive manner. Each block contains one $3\times 3$ deconv layer followed by one stride-1 $3\times 3$ conv. The final embedding map has the same spatial resolution as the input with depth equals to embedding dimension. In this way the decoder combines both high level contextual information as well as low-level details. }

\paragraph{Compression Module:}
\shenlong{We highlight the task-specific map compression module as the core contribution of this paper.}
The purpose of this module is to convert the large-resolution, high-precision embedding into a low-precision, lower-resolution one, without losing critical information for matching.
We employ  a fully convolutional encoder-decoder network to achieve this goal.
\shenlong{The neural network encoder is a fully convolutional residual network where each scale have two $3\times3$ standard residual blocks \cite{resnet} and a stride-2 $3\times 3$ conv between scales. The dimensionality per scale is $8, 16, 32, 64$ respectively. } 
The decoder is a fully convolutional network with several transposed convolutional layers. 
We use the PReLU~\cite{prelu} as the activation function. 
The output of the encoding module is passed through a grouped soft-max module, with a binarization module defined as 
\begin{equation}
\label{equ:binary}
p_j = \frac{\exp(f_j)}{\sum_{k \in \mathcal{S}_j} \exp(f_k)}, \,\,\,\,\,\, b_j = 
\left\{ \begin{array}{cc} 
1 & \text{\ if $p_j \geq 0.5$} \\
0 & \text{\ otherwise}
\end{array}\right.
\end{equation}
where $\mathcal{S}_j$ is the index group that j belongs to, with each group representing a non-overlapping subset of the full index set $\{ 1, \dots, K \}$; $\mathbf{f} = [f_0, \dots, f_i, \dots]$ is the input feature.  
The benefit of using grouped-softmax as encoder activation along with the binarizer is twofold.
First, within each group, we have at most  one non-zero entry. 
Thus, with the same number of channels it has better sparsity than the sigmoid function, increasing the compressibility of the binary encoding. 
Second, compared against standard soft-max, it increases the potential capacity since the grouping of indices allows a more structured encoding. \shenlong{While the component is non-differentiable, backpropagation was still feasible thanks to the use of a straight-through estimator, which we will show in Sec.~\ref{sec:learning} in detail.} \todo{I still think we need to mention this once since it's too far away to learning section. Don't want reader to bring this question until they reach the learning section.}

Thus, we only need to store onboard these highly compressed binary map
embeddings for localization. A two-step lossless binary encoding scheme is
adopted. Our first step is a Huffman encoding. The motivation is that the
frequency of appearance of items are not equal. The Huffman dictionary is built
by the one-hot encoding of the softmax latent probability $\bp$ per pixel.
Thus, for a 128-softmax vector the dictionary size is 128. Frequency is
computed in a batch manner. For instance, if the `class' $5 (00000101)$ appears
$50\%$ we could use a shorter-length code $0$ to encode it.  After that
a run-length encoding (RLE) is conducted over the flatten Huffman code map to
further reduce the size by making use of the fact that codes appear
consecutively. For instance $5555558$ could be further reduced to $5681$.  This
give us the final binary code that we store. Note that both Huffman encoding
and run-length encoding  are lossless. \shenlong{We choose Huffman+RLE due to
  its efficiency and effectiveness. Empirically, we also show that this
  approach reaches 72.5\% of
the ideal entropy lower bound. While other types of entropy coding, such as arithmetic coding exist, they are slower and bring marginal improvements to compression
rates~\cite{JPEGWiki}.}

Combining the NN encoding and the binary encoding, this full compressive encoding scheme gives a very large gain in terms of storage efficiency as shown later in the experimental section. 

\begin{figure}[t]
\vspace{-3mm}
\includegraphics[width=0.99\linewidth]{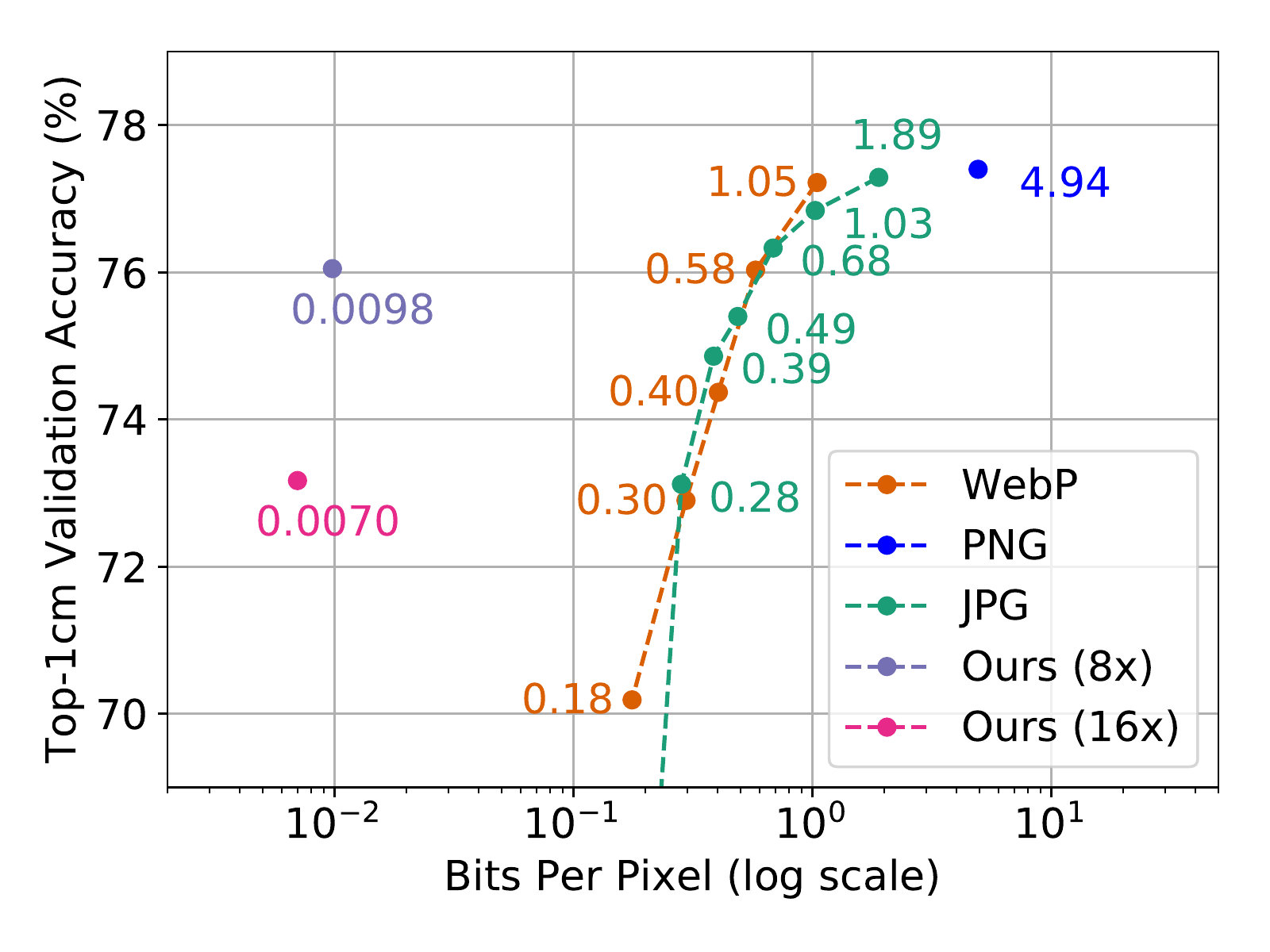} 
\vspace{-5mm}
   \caption{{\bf Top-1 Matching Performance vs.\ Bits per Pixel}}
\label{fig:matching_f}
\end{figure}

\begin{table}[]
\centering
\small
\begin{tabular}{lrrr}
\specialrule{.2em}{.1em}{.1em}
Method   & BPP & Top-9 px & Top-1 px \\ \midrule
Lossless (PNG)            & 4.93 & 97.47\% & \textbf{77.40\%} \\ %
Ours (recon, 8x)	 		  & 0.0520 & 97.27\% &  75.83\% \\
Ours (recon, 16x) 		  & 0.0140 & 96.95\% &  74.86\% \\ %
Ours (match, 8x)  & 0.0098 & \textbf{97.73\%} & 76.05\% \\ 
Ours (match, 16x) & \textbf{0.0070} & 97.25\% & 73.17\% \\ 
\specialrule{.2em}{.1em}{.1em}
\end{tabular}
\vspace{-3mm}
\caption{{\bf Ablation studies on matching performance.}
Optimizing jointly for both map reconstruction and matching greatly reduces the memory requirements.}
\label{tab:ab_matching}
\vspace{-3mm}
\end{table}

The decoder module then takes the binary code as input.
First, it transforms the Huffman+RLE codes back to the binary map, and
then applies a series of deconvolutional blocks to recover the full high resolution, high-precision  embedding of the map that we use for matching.
Fig.~\ref{fig:compression} illustrates the pipeline of the full compression module. 
\paragraph{Matching Module:}
Our matching module follows \cite{deep-gil}, where a  series of spatial transformer networks are utilized to rotate the online embedding multiple times at $|\Theta|$ different candidate angles.
Within each rotation angle, translational search based on inner-product similarity is equivalent to convolving the map embedding with the online embedding as kernels.
Thus, enumerating all the possible pose candidates is equivalent to a convolution with $|\Theta|$ kernels.
Unlike standard convolutions, this convolution has a very large kernel.
Following \cite{deep-gil}, we exploit FFT-conv to accelerate this matching modules by an order of magnitude (compared to GEMM-based convolutions) on a GPU. 

\subsection{End-to-End Learning}
\label{sec:learning}
\shenlong{Our full localization network is trained end-to-end, as the compression module is active during the training loop.}
Our loss function consists of two parts, namely  a matching loss that encourages that the end-task is accurate and a compression loss that minimizes the encoding length.
Thus, our total loss is defined as
\begin{align}
\ell =\ell_{\textsc{Loc}}(\mathbf{y}, \mathbf{y}_\textsc{gt}) + \lambda_1
\ell_{\textsc{CodeLen}}(\mathbf{p}) + \lambda_2 \ell_{\textsc{HardBin}}(\mathbf{p}),
\end{align}
where $\mathbf{y}$ is the final softmax-normalized matching score,   $ \mathbf{y}_\textsc{gt}$ is the one-hot representation of the ground truth (GT) position and  $\mathbf{p}$ is the embedding after the grouped-softmax layer in the compression module, defined in Eq.~\ref{equ:binary}. 

We employ cross-entropy as a matching loss.
This encourages the matching score to be the highest at the GT position, while lowering the score of positions elsewhere:
\[
\ell_{\textsc{Loc}}(\mathbf{y}, \mathbf{y}_\textsc{gt}) = \sum_i y_{\textsc{gt}, i} \log( y_i).
\]

The compression loss tries to minimize the encoding length. 
In particular, we use entropy as a differentiable surrogate of code length.
Note that this surrogate has been widely used in previous deep compression approaches \cite{toderici2015variable}.
According to Shannon's source coding theorem~\cite{shannon1948mathematical}, entropy provides an optimal code length, which could serve as a surrogate lower-bound for the actual encoding that we use.
Our entropy is estimated within each mini-batch as
\[
\ell_{\textsc{CodeLen}}(\mathbf{p}) =\bar{\bp} \log \bar{\bp},
\]
where $\bar{\mathbf{p}} = \frac{1}{W\times H\times B}\sum_i \bp_i$ is the mean soft-max probability averaged across all pixels' softmax probability $\bp_i$ in one batch example, defined as in Eq.~\ref{equ:binary}. In practice we find that this theoretical lower-bound is very close to the actual bit per pixel rate obtained after Huffman+RLE encoding.

Finally, we want the soft-max probability to be as close to one-hot as possible to reduce the loss due to hard binarization.
We thus  minimize each individual pixel's entropy as a regularization term:
\[
\ell_{\textsc{HardBin}}(\mathbf{p}) = \sum_i p_i \log p_i%
\]

Note that direct backpropagation is not feasible in our case,  as the binarization module defined in Eq.~\ref{equ:binary} is not differentiable. 
To overcome this issue, we adopt the straight-through estimator proposed in \cite{bengio2013estimating}.
That is, during the forward pass we conduct hard binarization, while during the backward pass we substitute this module with an identity function.
We find that this approximation provides good gradients for the function to be learned.

\subsection{Efficient Inference}

In the offline map encoding stage, we use our compression network to compress the map into a binary code such that the onboard storage requirements are minimized.
During onboard inference, the compressed code is recovered, the decoder is then used to create the HD embedding map, which is used for localization.

\begin{figure*}
 \centering 
 \begin{subfigure}{\textwidth}
   \centering
   \includegraphics[width=0.95\linewidth]{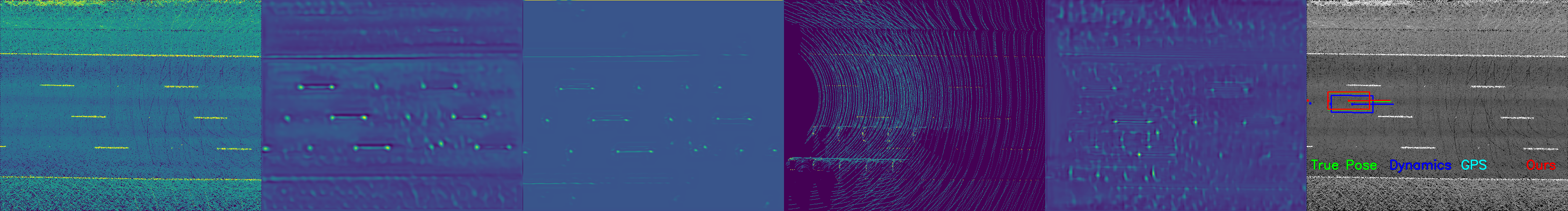}
 \end{subfigure}
 \begin{subfigure}{\textwidth}
   \centering
   \includegraphics[width=0.95\linewidth]{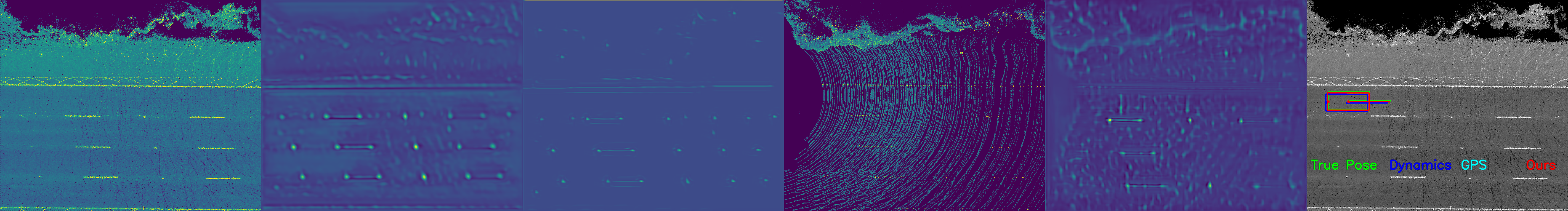}
 \end{subfigure}
 \vspace{-3mm}
 \caption{{\bf Qualitative results from our highway dataset.} From left to
    right: (1) original map, (2) its computed deep embedding, (3) the compressed 
    embedding, (4) online LiDAR observation, (5) its embedding, and (6) the
    localization result.}
\label{fig:loc_quali}
\end{figure*}
\paragraph{Onboard Inference:} 
Computing the exact probability defined in Eq.~\ref{eq:deepGIL} is not feasible due to the continuous space and the infeasible integral for the vehicle dynamics model defined in Eq.~\ref{eq:motion-model}. 
Following \cite{deep-gil}, we use a histogram filter to approximate the inference process. Towards this goal, we discretize the search space around a local region to a {5~$\times$~5~cm} grid. The integration will only be computed within this local trust-region, which neglects the rest of the solution space where the belief is negligible. In this manner, both the GPS and the dynamics term can be computed efficiently over a local grid as the search space. %
Unlike \cite{deep-gil}, when computing the LiDAR matching term $P_{\mlidar}(\cI_t | \bx; \bw)$, we first retrieve a local map binary code $\mathbf{b}$. The LiDAR embedding is computed through the feature network and then $\mathbf{b}$ is passed through the decoder of the compressor to recover the map embedding $g(\mathcal{M})$. After that, the matching score $P_\mlidar$ can be efficiently computed for all hypothesized poses as
a feed-forward network through FFT-conv. After each term has been computed, the final pose estimation is a soft-argmax aggregation taking uncertainty into consideration:
\begin{equation}
\label{eq:soft-argmax}
\bx_t^\ast= \frac{\sum_\bx \mathrm{Bel}_t(\bx)^\alpha \cdot \bx}{\sum_\bx \mathrm{Bel}_t(\bx)^\alpha}
\end{equation}
where $\alpha \geq 1$ is a temperature hyper-parameter.

\begin{table*}[]
\centering
\small
\begin{tabular}{lrrrrrrr}
\specialrule{.2em}{.1em}{.1em}
\textbf{Method}         & \multicolumn{3}{c}{\textbf{Median error (cm)}} & \multicolumn{3}{c}{\textbf{Failure rate (\%)}} & \textbf{Bit per pixel} \\
               & \multicolumn{1}{c}{Lat}    & Lon    & \multicolumn{1}{c}{Total}   & $\leq$ 100m & $\leq$ 500m & \multicolumn{1}{c}{End} &               \\  \midrule
    Lossless (PNG)       & \textbf{1.55} & \textbf{2.05} & \textbf{3.09} & \textbf{0.00} & \underline{1.09}       & 2.44          & 4.94            \\
JPG-50    & 3.29          & 5.60         & 7.59         & \textbf{0.00} & \underline{1.09}       & 5.26           & 1.03            \\
JPG-20    & 3.77          & 4.99          & 7.51          & \textbf{0.00} & \textbf{0.00} & 1.75          & 0.48            \\
JPG-10    & 3.42          & 5.46          & 7.54          & \textbf{0.00} & \underline{1.09}       &  5.26          & 0.28            \\
JPG-5     & 4.32          & 5.48           & 8.41           & \textbf{0.00}
          & \underline{1.09} & \underline{1.25}          & \underline{0.18}            \\
WebP-50   & 1.62           & 2.75          & 3.76         & \textbf{0.00} &3.26       & 3.30         & 1.05            \\
WebP-20   & 1.86         & 2.85         & 4.10          & 4.08  & 8.70  & 14.63          & 0.58            \\
WebP-10   & 1.60	& \underline{2.26}    & 3.48        & \textbf{0.00} & \underline{1.09}  & 2.50  & 0.40            \\
WebP-5    & 1.65	 & 5.75	& 6.53	& \underline{2.04}	& 5.43	& 13.95           & 0.30            \\
Ours & \underline{1.61}          & \underline{2.26}          & \underline{3.47}          & \textbf{0.00} & \underline{1.09}       & \textbf{1.22} & \textbf{0.0083} \\
\specialrule{.2em}{.1em}{.1em}
\end{tabular}
\vspace{-3mm}
\caption{{\bf Online localization performance on the urban dataset.}}
\label{tab:loc_urban}
\vspace{-3mm}
\end{table*}

\begin{table*}[ht]
\centering
\small
\begin{tabular}{lrrrrrrr}
\specialrule{.2em}{.1em}{.1em}
\textbf{Method}         & \multicolumn{3}{c}{\textbf{Median error (cm)}} & \multicolumn{3}{c}{\textbf{Failure rate (\%)}} & \textbf{Bit per pixel} \\
               & \multicolumn{1}{c}{Lat}    & Lon    & \multicolumn{1}{c}{Total}   & $\leq$ 100m & $\leq$ 500m & \multicolumn{1}{c}{End} &               \\  \midrule
Lossless (PNG)  & \textbf{3.62}       & \textbf{4.53}
                & \textbf{7.06}    & \textbf{0.00}        & \textbf{0.35}
                & \textbf{0.71}       & 4.94          \\ 
WebP-50        & \underline{3.87}         & 4.87            & 7.52     & \textbf{0.00}
               & \underline{0.71}        & \textbf{0.71}      & 1.46    \\ 
WebP-20        & 4.03         & 5.27            & 8.02     & \textbf{0.00}
               & 1.06        & \underline{8.87}       & 0.91    \\ 
WebP-10        & 4.45         & 7.09            & 9.79     & \underline{0.35}      & 9.57
               & 24.37       & 0.70    \\ 
WebP-5        & 4.10        & 6.40            & 8.99     & \underline{0.35}      & 9.57
              & 14.69       & \underline{0.55}   \\
Ours & \textbf{3.62}       & \underline{4.77}           & \underline{7.19}    & \underline{0.35}
      & \textbf{0.35}       & \textbf{0.71}       & \textbf{0.007}     \\
\specialrule{.2em}{.1em}{.1em}
\end{tabular}
\vspace{-3mm}
\caption{{\bf Online localization performance on the highway dataset.}}
\label{tab:loc_highway}
\end{table*}

\begin{table*}[]
\centering
\vspace{-2mm}
\small
\begin{tabular}{lrrrrrrr}
\specialrule{.2em}{.1em}{.1em}
\textbf{Method}         & \multicolumn{3}{c}{\textbf{Median error (cm)}} & \multicolumn{3}{c}{\textbf{Failure rate (\%)}} & \textbf{Bit per pixel} \\
               & \multicolumn{1}{c}{Lat}    & Lon    & \multicolumn{1}{c}{Total}   & $\leq$ 100m & $\leq$ 500m & \multicolumn{1}{c}{End} &               \\  \midrule

Lossless (PNG) & \textbf{1.55} & \textbf{2.05} & \textbf{3.09} &
\textbf{0.00} & \underline{1.09} & \underline{2.44} & 4.93580 \\
Ours (recon, 8$\times{}$)   &  \underline{1.59} & \underline{2.16}
                            & \underline{3.24} & \textbf{0.00}
                            & \underline{1.09} & \textbf{1.22} &  0.02689 \\
Ours (recon, 16$\times{}$)  & 1.76 & 2.48 & 3.62 & \textbf{0.00} & \textbf{0.00} & 2.56 &    0.01155   \\
Ours (match, 8$\times{}$)   & 1.61 & 2.26 & 3.47 & \textbf{0.00}
                            & \underline{1.09} & \textbf{1.22} &   \underline{0.00830} \\
Ours (match, 16$\times{}$)  &  1.62 & 2.77 & 3.84 & \underline{1.00} & 2.17 & 4.26 &  \textbf{0.00733} \\

\specialrule{.2em}{.1em}{.1em}
\end{tabular}
\vspace{-3mm}
\caption{{\bf Ablation studies on the urban dataset.}}
\label{tab:ab_urban}
\end{table*}

\section{Experimental Evaluation}
\vspace{-1mm}
\paragraph{Dataset:}
We evaluate our  approach over two large-scale driving datasets that cover highway and urban driving, respectively.
The highway dataset was  collected in \cite{deep-gil} and contains over 400 sequences of highway driving with a total of 3\,000 km travelled.
It contains an HD, dense, LiDAR intensity map stored in lossless PNG format.
The self-driving vehicle integrates a 64-line LiDAR sweeping at 10Hz, with GPS and IMU sensor.
We follow the setting of \cite{deep-gil} and select 282 km of driving as testing, ensuring that there is no geographic overlap between the splits.
The GT localization is estimated by a high-precision offline Graph-SLAM.

To better evaluate the potential of the model to compress maps
with more diverse content and complicated structures, we build a new urban
driving dataset that consists of {15\,554} km of driving. This dataset is
collected in a metropolitan city in North America with diverse scenes and road
structures.
This dataset is more challenging as it introduces more diverse
vehicle maneuvers, including sharp turns and reverse driving, as well as some regions with poor lane markings and map changes.
The ground-truth localization is estimated through an high-precision offline Graph-SLAM with multi-sensor fusion.
Intensity maps are built by multiple passes through a comprehensive offline pose graph optimization at a resolution of 5cm per pixel. 

\vspace{-3mm}
\paragraph{Experimental Setup:}
To our knowledge, no previous work has integrated LiDAR intensity localization with deep compression.
Therefore, %
we evaluate our work against baselines without compression on localization metrics alone, such as those found in \cite{deep-gil}, and measure the performance degradation when using the map compression module.

Since the intensity map is stored as an image,
we compare against several traditional image compression algorithms
such as JPEG and WebP.  
For each compression algorithm, we compress the training and testing map images and train a standard learn-to-localize matching network \cite{deep-gil}.
We also train a reconstruction-based compression network \shenlong{(`ours (recon)')} that shares the same architecture with our compression module, with the only exception that it is trained for reconstruction error of the feature map only (not for matching performance).
This showcases whether the task-specific compression helps our matching task.

\shenlong{For our proposed method, we adopt two different settings for compresssion, by changing the downsampling levels we used for our binary codes. We have 8x downsampling model and 16x downsampling model}, where the 16x model has an extra set of downsampling and upsampling modules before binarization, which varies the compression rate.
All the competing algorithms have the same embedding feature network as our proposed model. \shenlong{Additionally, we performed experiments with reduced map resolution as an alternative baseline for reducing map storage.}

We train all competing algorithms over 343k and 230k training samples for urban and highway datasets respectively.
We aggregate five online LiDAR sweeps and rasterize them into a birds' eye view image at 5cm/pixel, in ranges of {($-$12 m, 12 m) and ($-$15 m, 15 m)}.
All networks are trained on four NVIDIA 1080 Ti GPUs using PyTorch.
We use the Adam optimizer \cite{kingma2014adam} with an initial learning rate of $10^{-3}$.
We observed that training the entire network end-to-end from scratch works, but is slower to converge.
To speed up training, we first train the non-compressed matching network  without our additions for the localization task, and then insert our compression module and train end-to-end.   

\vspace{-3mm}
\paragraph{Matching Performance:}
In order to evaluate the performance of the models in terms of finding the best match in a compressed map, we report the performance of the competing algorithms under the matching setting.

We conduct matching over a $1$ m\textsuperscript{2} search range, after perturbing the initial position of the vehicle around the GT position.
We sample the translational perturbation between 0 and $1$ m\textsuperscript{2},
and the angular perturbation between 0 and 5$^\circ$, both uniformly.
We report top-1 px and top-9 px as our metrics, representing whether the prediction is in the same pixel as the GT or within the $3\times3$ region centered around the GT, respectively.
We report matching accuracy as a function of bit rate per pixel on the urban dataset in Fig.~\ref{fig:matching_f}. 
Note that the proposed algorithm at 8x setting achieves 76\% top-1 px accuracy with 0.0098 bit per pixel rate.
Both are higher than all competing algorithms.
Also, it obtains similar top-9 px accuracy on par with no compression module.
Especially, the BPP is around 20-400 times smaller than all competing algorithms. Under the 16x setting the top 1 px accuracy is 3\% lower but achieves a higher compression rate at 0.007 bits per pixel.

\vspace{-3mm}
\paragraph{Ablation Studies:}
We conducted an ablation study over the matching performance.
We first validate whether jointly training the compression module with our matching task loss helps improve the matching performance and increase the compression rate.
For this, we train a compression module using only reconstruction loss (without the matching task loss).
Secondly, we report whether a lower compression rate is achieved through aggressive map downsampling.  Table~\ref{tab:ab_matching} illustrates the results.
We can see that jointly training the compression module with the task specific loss greatly helps the performance.
The 16x downsampled model pushes the compression rate even further, with a 3\% percent drop on top 1px results.

\vspace{-3mm}
\paragraph{Online Localization:}
We follow \cite{deep-gil} and 
compute the median and worst case localization error on the test split as our metrics.
To be specific, we report median, p95, and p99 error in meters along the lateral and longitudinal directions.
We also report an out-of-range rate, which represents the percentage of
1 km segments where the method reaches a localization error of 1m.

Table~\ref{tab:loc_urban} shows the online localization performance on the urban dataset.
While most of the baselines provide reasonable results,
our method is clearly better than competing algorithms such as JPG-5 and WebP-5, which show high failure rates at extreme compression levels.
In terms of worst case, measured by failure rate, our method is on par with high-quality compression such as WebP-50 and JPEG-50, and a lossless method, while our bit rate per pixel is 100 times smaller. This is shown in in Fig.~\ref{fig:failure_f}, where we plot the percentage of failures after 1 km against the storage.

Table~\ref{tab:loc_highway} depicts the online localization performance on the highway dataset.
We can see that traditional off-the-shelf compression algorithms like WebP have a large performance drop compared to our compression-based matching. %
While the method using reconstruction loss obtains storage roughly in the same magnitude as our approach, it suffers a large performance drop.
This indicates the importance of matching loss term for effectively selecting portions of the map to keep. Meanwhile, our method based on the matching task loss has  no performance drop at half the storage of the pure reconstruction network, nor at more than 400 times smaller compared to the lossless compression bitrate.

Table~\ref{tab:ab_urban} showcases the ablation study on the urban dataset.
We compare reconstruction loss driven compression models against our matching-loss driven compression model under various architectures.
From the Tables we can see that, in terms of online localization error, our compression model trained with task-specific driven loss is better than the reconstruction model with %
smaller bitrates and a lower failure rate. 

\vspace{-3mm}
\paragraph{Qualitative Analysis:}
Fig.~\ref{fig:loc_quali} shows examples of the deep map embeddings computed by
our system (before and after compression) together with the (uncompressed)
online observation embedding and the localization result. For more
results please refer to the supplementary material.

\vspace{-3mm}
\paragraph{Storage Analysis:}
We now turn back to the approximate storage
requirements described in the introduction, and showcase projected numbers when
compressing all maps using our proposed method in
Tab.~\ref{tab:storage-analysis}. Our proposed method can compress a 5cm/px 
HD map of the entire Los Angeles county to just 1.5 GB, allowing it to fit in
RAM on most current smartphones. 
We can also fit the entire USA road network at the same resolution in just 280
GB.

\begin{table}
\vspace{-3mm}
  \centering
  \small
  \begin{tabular}{lrr}
\specialrule{.2em}{.1em}{.1em}
    \textbf{Compression} & \textbf{LA County} & \textbf{Full US} \\
    \midrule
    Lossless (PNG)  & 900 GB & 168 TB \\
    WebP 1 & 32 GB  & 5.99 TB \\ 
    Ours (match, 8x) & \textbf{1.5 GB}  & \textbf{0.28 TB} \\
\specialrule{.2em}{.1em}{.1em}
  \end{tabular}
  \vspace{-3mm}
  \caption{{\bf Estimated map storage requirements using various compression
    methods.\label{tab:storage-analysis}}}
  \vspace{-3mm}
\end{table}

\section{Conclusions}

HD maps impose high storage requirements, which limit
the ability of a self-driving fleet
to operate in large-scale environments.
In this paper, we proposed to learn to compress the map
representation such that it is optimal for the localization task.
Our experiments on a state-of-the-art localizer have shown that it is possible to learn a task-specific compression scheme  that
reduces storage requirements by two orders of magnitude compared to general-purpose
codecs such as WebP, without sacrificing localization performance.

{\small
\bibliographystyle{ieee}
\bibliography{egbib}
}

\onecolumn
\appendix
\section{Additional Experimental Results}

\subsection{Quantitative Results}

\paragraph{Comparison to smaller map resolutions:} 

We also performed experiments with reduced map resolutions on our urban
dataset to investigate the impact on storage requirements and localization
accuracy.
As shown in Table~\ref{tab:lowres},
We note that unlike the tables in the
paper, here we measure the storage requirements in bits / m$^2$, in order to
account for the different map resolutions. 
magnitude of the storage required by our approach. However,  the localization performance
is substantially reduced (16.28\% failure rate, as opposed to 2.56\% for our
binary coding).

\newcommand{\second}[1]{\underline{#1}}

\begin{table*}[]
\centering
\begin{tabular}{lccccccc}
  \toprule
\textbf{Method}         & \multicolumn{3}{c}{\textbf{Median Err (cm)}}
                        & \multicolumn{3}{c}{\textbf{Failure Rate (\%)}}
                        & \textbf{b/m$^2$} \\
               & \multicolumn{1}{c}{Lat}    & Lon
               & \multicolumn{1}{c}{Total}   & $\leq$ 100m & $\leq$500m
               & \multicolumn{1}{c}{End} &               \\  \midrule
%
               PNG, 5cm/px &
               \textbf{1.55} & \textbf{2.05} &	\textbf{3.09} & \textbf{0.00} &
               \second{1.09} & \textbf{2.44}
               & 1948.55          \\ 

PNG, 10cm/px & 4.37 & 6.68 & 9.50 & 3.19 & 3.26 & 4.00 & 402.84 \\
JPG@50, 10cm/px & 4.51 & 5.78 & 8.95 & 0.00 & \second{1.09} & 10.64 & 63.42 \\

PNG, 15cm/px & 15.73 & 23.66 & 31.73 & 10.31 & 20.65 & 22.03 & 173.97 \\
JPG@50, 15cm/px & 11.67 &  18.20 & 25.14 & 9.28 & 13.04 & 16.28 & \second{29.00} \\

%
Ours (16$\times{}$) & \second{1.76}  &	\second{2.48} & 	\second{3.62} &
\textbf{0.00} & \textbf{0.00} & \second{2.56} & \textbf{2.87} \\

\bottomrule
\end{tabular}
\caption{Localization performance on our urban dataset using reduced
resolution maps. We used 5cm/px in the submission.
Map storage is measured in bits/m$^2$ in order
to account for different resolutions (bits-per-pixel (bpp) are no longer meaningful
if the area of a pixel can change).
\emph{Ours} refers to our 16$\times$ downsampling
method. JPG quality is 50.}
\label{tab:lowres}
\end{table*}

\subsection{Qualitative Results}

%
\newcommand{\includetrim}[5]{
  \adjincludegraphics[width=\linewidth, trim={{#1\width} {#2\height} {#3\width} {#4\height}},clip,valign=T]{#5}
}

\newcolumntype{P}[1]{>{\centering\arraybackslash}p{#1}}
\newcolumntype{L}[1]{>{}p{#1}}

\newcounter{locfigno}

\newcommand{\QualFig}[2]{

\def\rootbase{../figures/matching_samples/000-torloc_24batch_14_24_notfullyconverge62k_viz_only_compress_binarize_8x_5e-2}

\begin{figure}
  \centering
    \begin{subfigure}[c]{0.45\textwidth}
    \begin{center}
        \includegraphics[width=\linewidth]{\rootbase_Input_Map_#1}
    \end{center}
    \vspace{-0.25cm}
    \caption{Input Prior Map}
    \end{subfigure}
    \begin{subfigure}[c]{0.45\textwidth}
    \begin{center}
        \includegraphics[width=\linewidth]{\rootbase_Input_Online_#1}
    \end{center}
    \vspace{-0.25cm}
    \caption{Input LiDAR Observation}
    \end{subfigure}
    \begin{subfigure}[c]{0.45\textwidth}
    \begin{center}
        \includegraphics[width=\linewidth]{\rootbase_cbc_viz_IntermediateMap_#1}
    \end{center}
    \vspace{-0.25cm}
    \caption{Uncompressed Map Embedding}
    \end{subfigure}
    \begin{subfigure}[c]{0.45\textwidth}
    \begin{center}
        \includegraphics[width=\linewidth]{\rootbase_Embedding_Online_#1}
    \end{center}
    \vspace{-0.25cm}
    \caption{Online Embedding}
    \vspace{0.25cm}
    \end{subfigure}
    \begin{subfigure}[c]{0.45\textwidth}
    \begin{center}
        \includegraphics[width=\linewidth]{\rootbase_Embedding_Map_#1}
    \end{center}
    \vspace{-0.25cm}
    \caption{Compressed Map Embedding}
    \end{subfigure}

  \caption{#2 \label{fig:loc-qual-\arabic{locfigno}}}
  \stepcounter{locfigno}
\end{figure}
}   

\newcommand{\QualFigBin}[2]{

\def\rootbase{../figures/matching_samples/000-torloc_24batch_14_24_notfullyconverge62k_viz_only_compress_binarize_8x_5e-2}

\begin{figure}
  \centering
  \begin{minipage}{0.5\linewidth}
    \begin{subfigure}[c]{0.90\textwidth}
    \begin{center}
        \includegraphics[width=\linewidth]{\rootbase_Input_Map_#1}
    \end{center}
    \vspace{-0.25cm}
    \caption{Input Prior Map}
    \end{subfigure}
    \begin{subfigure}[c]{0.90\textwidth}
    \begin{center}
        \includegraphics[width=\linewidth]{\rootbase_cbc_viz_IntermediateMap_#1}
    \end{center}
    \vspace{-0.25cm}
    \caption{Uncompressed Map Embedding}
    \end{subfigure}
    \begin{subfigure}[c]{0.90\textwidth}
    \begin{center}
        \includegraphics[width=\linewidth]{\rootbase_Embedding_Map_#1}
    \end{center}
    \vspace{-0.25cm}
    \caption{Compressed Map Embedding (i.e., reconstructed from the binary codes)}
    \end{subfigure}
  \end{minipage}
    \begin{subfigure}[c]{0.373\textwidth}
    \begin{center}
        \includegraphics[width=\linewidth]{\rootbase_cbc_viz_BinaryEmbedding_#1}
    \end{center}
    \vspace{-0.25cm}
    \caption{Binary codes learned by our system. Their sparsity makes them very
    easy to compress.}
    \end{subfigure}

    \caption{#2 \label{fig:bin-code}}
\end{figure}
}   

\QualFig{31600}{Preview of the localizer operating in a regular intersection.}
\QualFig{33200}{Example where the online localization successfully deals with
heavy traffic.}
\QualFig{28600}{Example of a side road with no lane markings.}
\QualFig{29500}{A section of the map with tram lines.}
\QualFig{34700}{A parking lot.}
\QualFig{41600}{An intersection with fainter-than-usual crosswalks.}
\QualFig{43200}{Unusual crosswalks.}

\QualFigBin{30600}{Examples of inputs to our system, together with the computed
binary codes used to represent the learned map embedding in a compact way.
Recall that the binary code maps are lower resolution than the inputs (this
example uses the $8\times{}$ downsampling, so each code has $1/8$ the
resolution of the input). Moreover, the neural network learns to only use
a limited subset of the possible binary codes, leading to the reduced storage
requirements described in the experimental section.}

\addtocounter{locfigno}{-1}

Figures~\ref{fig:loc-qual-0}--\ref{fig:loc-qual-\arabic{locfigno}} contain
samples from our localization application. Note that the compression happens
independently of the online observations. Here, we show the online observations
and their embedding for reference, and to highlight that our matching is robust
to any traffic conditions.

We note that in all our visualizations the original map is shown for
illustration purposes only. In practice, the original map does not need to be
saved onboard, as the compressed embedding is enough for performing online
localization.

Figure~\ref{fig:bin-code} shows an example of the binary codes generated by
our compression module.
This example shows the 64-way binarized embedding that is the
output of our grouped softmax compression module. We note that only some of the
channels are activated, and thus most of the binary bits correspond to a small
subset of the binary embedding channels.  This is a result of having an
optimization objective that consists of entropy-based losses, and these sparse
results show that we successfully learned to discard unused channels, keeping
channels only if they capture information important for our localization task.
We observe that the compression scheme learns to dedicate channels to represent
important geometric features such as road boundaries, and lane markings. 

For further results, we would like to refer the reader to the video associated
with this submission, which shows our probabilistic localizer running online
using a compressed map embedding.


\end{document}